\documentclass[10pt,letterpaper]{article}
\pdfoutput=1

\usepackage{cogsci}

\cogscifinalcopy 

\usepackage{graphicx}
\usepackage{hyperref}
\usepackage{xcolor}
\usepackage{colortbl}
\usepackage{soul}
\usepackage{amsfonts}
\usepackage{stmaryrd}
\usepackage[T1]{fontenc}
\usepackage{enumitem}
\usepackage{hyperref}
\usepackage{afterpage}
\usepackage{amsmath}
\usepackage{amssymb}
\usepackage{amsthm}
\usepackage{multirow} 
\usepackage{balance}
\usepackage{subcaption}
\usepackage{stfloats}
\usepackage{rotating}
\usepackage{makecell}
\usepackage{hhline}
\usepackage{xspace}
\usepackage{algorithm}
\usepackage{algpseudocode}
\usepackage{cancel}
\usepackage{booktabs}
\usepackage{mathrsfs}  
\usepackage{tipa}
\usepackage{wrapfig}

\usepackage{pslatex}
\usepackage{apacite}
\usepackage{float}

\title{The Greedy and Recursive Search for Morphological Productivity}

\author{
Caleb Belth$^\mathbf{1}$ (cbelth@umich.edu), 
{\bf Sarah Payne$^\mathbf{2}$ (paynesa@sas.upenn.edu)}, 
{\bf Deniz Beser$^\mathbf{3}$ (beser@isi.edu)},\\ 
{\bf Jordan Kodner$^\mathbf{4,5}$ (jordan.kodner@stonybrook.edu)}, 
{\bf Charles Yang$^{\mathbf{2,6}}$ (charles.yang@ling.upenn.edu)} \\
{$^1$Department of Computer Science and Engineering, University of Michigan} \\
{$^2$Department of Linguistics and Department of Computer and Information Science, University of Pennsylvania} \\
{$^3$Information Sciences Institute, University of Southern California}\\
{$^4$Department of Linguistics, Stony Brook University}\\
{$^5$Institute for Advanced Computational Science, Stony Brook University}\\
{$^6$Department of Psychology, University of Pennsylvania} \\
}

\definecolor{darkspringgreen}{rgb}{0.09, 0.45, 0.27}
\definecolor{darkred}{rgb}{0.55, 0.0, 0.0}

\definecolor{alizarin}{rgb}{0.82, 0.1, 0.26}
\definecolor{ao(english)}{rgb}{0.0, 0.5, 0.0}
\definecolor{cadmiumgreen}{rgb}{0.0, 0.42, 0.24}

\DeclareMathOperator*{\argmax}{arg\,max}

\newcommand{\method}{\textsc{ATP}\xspace}
\newcommand{\tp}{\textsc{TP}\xspace}

\newcommand{\childes}{\textsc{CHILDES}\xspace}

\newcommand{\celexEng}{\textsc{CELEX-EN}\xspace}
\newcommand{\childesEng}{\textsc{CHILDES-EN}\xspace}
\newcommand{\childesGer}{\textsc{CHILDES-DE}\xspace}

\newcommand{\lemma}{\ell}
\newcommand{\inflected}{\textsc{i}}
\newcommand{\feats}{\mathcal{F}}

\newcommand{\R}{r}

\newcommand{\featureSpace}{\Omega}
\newcommand{\splitFeature}{s}

\newcommand{\exceptions}{\emph{e}}
\newcommand{\num}{\emph{N}}
\newcommand{\threshold}{\theta}

\newcommand{\maxSuffix}{\sigma_{\max}}
\newcommand{\lemmaEndings}{\mathcal{E}}
\newcommand{\haveSuffix}{\mathcal{S}}
\newcommand{\freq}{f}

\newcommand{\train}{\mathcal{X}}

\newcommand{\ipaI}{\textsci}

\newcommand{\citet}[1]{\citeA{#1}}

\begin{document}

\maketitle

\begin{abstract}
As children acquire the knowledge of their language's morphology, they invariably discover the productive processes that can generalize to new words. Morphological learning is made challenging by the fact that even fully productive rules have exceptions, as in the well-known case of English past tense verbs, which features the -\textit{ed} rule against the irregular verbs. The Tolerance Principle is a recent proposal that provides a precise threshold of exceptions that a productive rule can withstand. Its empirical application so far, however, requires the researcher to fully specify rules defined over a set of words. We propose a greedy search model that automatically hypothesizes rules and evaluates their productivity over a vocabulary. When the search for broader productivity fails, the model recursively subdivides the vocabulary and continues the search for productivity over narrower rules. Trained on psychologically realistic data from child-directed input, our model displays developmental patterns observed in child morphology acquisition, including the notoriously complex case of German noun pluralization. It also produces responses to nonce words that, despite receiving only a fraction of the training data, are more similar to those of human subjects than current neural network models' responses are.

\end{abstract}

\section{Introduction}
The acquisition of English past tense is one of the most extensively studied problems in cognitive science \cite{McClelland2002, Pinker2002}. Yet its simplicity---a single, and numerically dominant, rule (-\textit{ed}) along with a list of irregular verb exceptions---is hardly representative of the complexity of the world's morphological systems \cite{comrie1989language}. In comparison, German noun pluralization is a more challenging test for morphological learning theories. The plural is formed by five suffixes, but even the \emph{least} frequent, -\textit{s}, is productive and  applies to novel nouns (e.g., {\it iPhones}), while the other four suffixes are also productive for sub-categories of nouns characterized by their gender and phonological properties \cite{Wiese1996}.   Nevertheless, children exhibit remarkable proficiency of their native language's morphology at a very early age; see \cite{Lignos2016} for a cross-linguistic review. 

Children's discovery of productive rules is also attested experimentally when presenting young children with nonce words in a Wug test \cite{berko1958child} and when new words (e.g., \emph{google-googled}) enter a language. As a result, traditional linguistic approaches to language acquisition often make use of productive rules \cite{SPE}. 
\citet{rumelhart1986learning} proposed an impactful neural-network (NN) model, suggesting that NNs might be capable of exhibiting rule-like behavior despite having no explicit representation of rules. Ensuing critiques of the model by \citet{pinker1988language} and of connectionist models of cognition more generally by \citet{fodor1988connectionism} sparked what came to be known as the ``Past Tense Debate.''
Due to the extensive advancements in NN architecture in natural language processing and machine learning, the debate has recently been revived by \citet{Kirov2018RecurrentNN} with some initially promising improvements over early models (in particular, more realistic accuracy). However, these accomplishments have been challenged, especially when compared against human behavioral results on Wug tests in English past tense  \cite{corkery2019we9} and German plurals \cite{McCurdy2020InflectingWT}. 

The Tolerance Principle (\tp) may provide another solution to some of these challenges \cite{yang2016price}. It provides a tipping point for rule productivity based on the level of rote-memorized exceptions that the rule needs to withstand. 
A mathematical consequence of the TP is that rules defined over smaller vocabularies can tolerate a larger fraction of exceptions. This property is attractive. First, it makes the TP a promising candidate for modeling early language development, when children's vocabulary size is quite limited. Second, if the learner fails to acquire a broad rule defined over a vocabulary, the vocabulary can be partitioned into subsets, within which narrower rules may be identified by recursive application of the \tp. 
In this work, we propose an abductive learning model of morphology, which we call \emph{Abduction of Tolerable Productivity} (\method). \method is a search procedure that recursively hypothesizes rules and evaluates productivity, until all words in the vocabulary are either accounted for by productive rules or listed as exceptions.

\section{Background}
\subsection{Psychological and Developmental Considerations}
Computational models of cognition, which inevitably make simplifying assumptions, must still operate within the boundaries established by empirical research. We review several important lines of results from child language acquisition, which serve as the design specifications for computational models of morphological learning. 

\subsubsection{Data size} One of the most remarkable characteristics of child language acquisition is how small the vocabulary on which they learn is. All English-learning children acquire the productive use of  past tense by age three \cite{Kuczaj1977}, some even before age two  \cite{Brown1973}. 
Likewise, German-learners show over-regularization of suffixes such as -\textit{(e)n} and -\textit{e} before or around age two \cite{Mills1986, Elsen2002}.  At such an early stage of development, children have a very modest vocabulary: a two-year-old English learner's total vocabulary is 500 at most, and a three-year-old's vocabulary has an upper limit of just over 1000; most children's vocabulary size is considerably smaller \cite{Fenson1994, Hart1995}. See \cite{Bornstein2004, Szagun2006} for similar findings in other languages.  Thus, a developmentally plausible cognitive model should be able to learn the core morphology of inflection from datasets containing only a few hundred words.

\subsubsection{Productivity and Wug Tests} 
The classic study by \citet{berko1958child} was the first systematic demonstration that young children apply productive rules to form noun plurals ({\it wug-wugs}) and past tense verbs ({\it rick-ricked}). Berko also found that children nearly categorically resist the analogical use of irregular forms, even for words very similar to existing irregulars. For example, only 1 of the 86 children in the study produced {\it glang} for the novel verb {\it gling}. The categorical status of productivity is also strongly confirmed in naturalistic production. One of the first quantitative longitudinal studies \cite{MacWhinney1978} noted that regular forms are rarely analogized into irregular forms, while irregular forms are often regularized.  More recent quantitative studies estimate the rate of past tense over-regularization at 8-10\% of all past tense forms \cite{Maratsos2000, Maslen2004}. By contrast, children almost never over-extend an irregular form (e.g., {\it bite-bote} from {\it write-wrote}, {\it fry-frew} from {\it fly-flew}): the most comprehensive study places the error rate at 0.2\% \cite{Xu1995}.

\subsection{The Tolerance Principle}
\label{subsec:tp}

The Tolerance Principle (\tp) is a cognitively-motivated, theoretical tipping point that makes quantifiable predictions about when a linguistic process is used productively. It is inspired by studies of lexical processing and the hypothesis that children use a morphological process productively when it is computationally more efficient to do so. The \tp only depends on two quantities: $\num$---the number of words in the rule's scope---and $\exceptions$, the number of exceptions. For instance, the -s process for English pluralization applies to singular nouns and has exceptions like \emph{children, sheep, fish}, etc. In our model, rules are of the form $\R: A \implies C$ where $A$ is the antecedent and $C$ the consequent; $\num$ measures how many times $A$ applies and $\exceptions$ measures the number of times $C$ fails to follow from $A$.
Given a rule $\R$ with a scope of size $\num$ and $\exceptions$ exceptions, the \tp states that
\begin{equation}
    \R \text{ is productive iff } \exceptions \leq \threshold_\num \triangleq \frac{\num}{\ln{\num}}.
\end{equation}
\noindent
The \tp has consistently made accurate predictions on when children accept rules as productive and when they do not, as
confirmed in artificial language learning experiments \cite{schuler2016testing, Koulaguina2019} with precisely controlled conditions. Productivity under the \tp is  categorical, which mirrors children's morphological use reviewed above. It is also parameter-free:  the two values  $\num$ and $\exceptions$ are word counts directly from the training data that require neither parameter tuning nor statistical fitting. 

Critically for complex morphological systems, the \tp can be applied recursively: if a rule $\R$ is unproductive over its current scope, its scope can be recursively narrowed. For example, none of the five German noun plural suffixes covers a sufficiently large number of nouns to tolerate the rest as exceptions: the learner will attempt to organize nouns into subcategories---defined by morphological gender and the phonological form of the noun---to recursively search for productivity within. As we will see, this divide-and-conquer strategy is fully automated in our search procedure.

\section{Model}
\subsection{Abductive Search for Productivity}

The recursive application of the \tp lends itself to a Peircean abductive learning procedure. Given data in pairs, such a procedure hypothesizes rules that map one set to the other (e.g., lemmas to their inflected forms). If no such rule is productive via the \tp, then it subdivides the words according to some feature---that is, it refines the hypothesis over more narrow scopes---and recursively tries again.

We propose just such a procedure: \emph{Abduction of Tolerable Productivity} (\method).
Its input is a set $\train$ of \textbf{instances} in the form ($\lemma, \feats, \inflected$), where $\lemma$ is a lemma, $\feats \subseteq \featureSpace$ is a set of features from the \textbf{feature space} $\featureSpace$, and $\inflected$ is the inflected form corresponding to lemma $\lemma$ and features $\feats$; for instance (\texttt{walk}, $\{$\texttt{3,SG,PST}$\}$, \texttt{walked}), where the example features \texttt{3,SG,PST} carry the information that the inflection is 3rd person, singular, and past tense.

\method recursively grows a decision tree, where each instance's ``label'' is the morphological change that produces the inflection; the resulting tree thus encodes a map from lemma and features to inflection. In principle \method could model any type of inflectional morphology, but the inflections modeled in this paper involve only suffixation, in which case an instance's ``label'' is the suffix that is concatenated to the lemma. For clarity, we describe \method in terms of suffixation.

At each recursive level, 
the decision of which feature to split on is selected to maximize consistency: the relative frequency of the most frequent suffix that the instances with that feature take. That is, it splits on
$\hat{\splitFeature} = \argmax_{\splitFeature \in \featureSpace} = \frac{\freq_{\maxSuffix(\train_\splitFeature)}}{|\train_\splitFeature|}$, 
where $\train_\splitFeature = \{(\lemma, \feats, \inflected) \in \train :\splitFeature \in \feats\}$ is the set of instances with feature $\splitFeature$ and $\freq_{\maxSuffix(\train_\splitFeature)}$ is the frequency of the most frequent suffix in $\train_\splitFeature$. The split is formed by recursing separately on those instances with feature $\splitFeature$ (i.e., $\train_\splitFeature$) and those without it (i.e., $\train \setminus \train_\splitFeature$).

If \emph{productivity} is defined as the frequency of the most frequent suffix in $\train_\splitFeature$, then this is one possible formulation of \citet{yang2016price}'s proposed \emph{Principle of Maximize Productivity}, which he described as ``Pursue rules that maximize productivity.'' Furthermore, it is motivated by findings that consistent patterns promote generalization and category formation in learning \cite{Gerken2006, Reeder2013}. 
\vspace{0.15cm}
\subsubsection{Features}

The set of features provided as input is expanded to include regularities in the ending of lemmas.
For instance, the null suffix for German nouns is predominantly used for non-feminine lemmas ending with a schwa followed by \textit{l}, \textit{r}, or \textit{n} \cite{Wiese1996}. Regularities of this sort are extracted at each recursive level as the shortest lemma endings that productively predict a suffix in $\train$ (the training instances at that level). 

To do so, \method considers suffixes one-by-one and, for each suffix, it considers the endings of lemmas that take that suffix, starting with the shortest ending. It caches any ending where the number of lemmas that do have the ending but do \emph{not} take the suffix is tolerably low; effectively, \emph{ending} $\implies\ $ \emph{suffix} passes the TP. For example, enough English verb lemmas ending in [k] take the [-t] suffix to pass the TP; similarly for [p], [\textesh], and other voiceless segments. We denote the lemmas that take any such endings as $\lemmaEndings$ (e.g., $\lemmaEndings = $ \{lemma : lemma ends in [k] or [p] or ... or [\textesh]\}
in the English verb example). If $\haveSuffix$ consists of the lemmas whose inflections take the suffix being considered, then \method tests the TP for both $\num_1 = |\lemmaEndings|, \exceptions_1 = |\lemmaEndings \setminus \haveSuffix|$ and $\num_2 = |\haveSuffix|, \exceptions_2 = |\haveSuffix \setminus \lemmaEndings|$. If both tests pass the TP, the endings contributing to $\lemmaEndings$ are added to the feature space $\featureSpace$. The purpose of these tests is to establish that there is a productive relationship between the endings and the suffix. In the running example, all words that end in a voiceless segment other than [t] take the [-t] suffix, and all words that take the [-t] suffix end in a voiceless segment, so both tests pass.
Further examples of the result can be seen in Figs.~\ref{fig:english-tree}-\ref{fig:german-tree}, where sets of lemma endings appear in brackets, separated by ``|'' for \emph{logical or}. This has the interpretation of picking out the lemmas that end in any such ending (i.e., $\lemmaEndings$).

Thus, at the end of the lemma ending extraction, the feature space $\featureSpace$ consists of whatever features were provided as input, together with the newly added sets of endings.
\method is agnostic to the nature of a feature, such as whether it has to do with content or form. It considers them all equally, seeking to maximize consistency, as described above. 

Moreover, alternative approaches to incorporating or discovering features could be used without changing the fundamental recursive, abductive search procedure. Such adaptions could be useful in morphological processes beyond suffixation, where \method could make use of features relevant to processes such as infixation and stem change within the same abductive search procedure. 
\vspace{0.15cm}
\subsubsection{Recursive and Base Cases} 
Once a split is performed, the node's children are formed recursively on the partitioned set. Features that are equivalent across all instances in $\train$ are ignored since they are completely uninformative. The base case of this recursion is reached when the most frequent suffix at the node passes the \tp or when there are no more features to split on. The path to a node with a productive suffix is the rule (e.g., Figs.~\ref{fig:english-tree}-\ref{fig:german-tree}), and all exceptions to the rule are memorized by storing them at the corresponding node. If a node is reached where no suffix is productive and there are no more features to split on, the node's instances are memorized.

\subsection{Inflection Production}

Once trained, \method can use its acquired knowledge to produce the inflected form corresponding to a lemma and its features. It does so by traversing the decision tree to a leaf. If the leaf has a productive suffix, \method produces the inflected form via the rule. If there is no productive suffix at the leaf, \method makes an analogical guess by retrieving the memorized lemma at that node with the smallest character-level Hamming distance, padding 0's if necessary to make the two strings the same length. \method uses the learned inflected form for the nearest-neighbor lemma to produce the inflection for the target lemma. In some cases---such as when nonce words are presented---some features may be unknown. In this case, an inflection is produced by traversing all logically compatible paths and using the most specific (i.e., deepest in the tree) compatible rule. A path is \emph{logically compatible} if none of the features it specifies are contradicted by the word's known features. For instance, if the gender of a nonce word is unknown, and \method encounters a branch point in the tree that is determined by gender, it takes both branches. 

It is worth pointing out that analogical guessing forces an answer in order to quantitatively evaluate our model on test data. However, in many cases of language use, 
speakers may refuse to produce any form when they do not have an applicable productive rule, as in the well-known case of morphological gaps \cite{yang2016price}.

\subsubsection{Code} We make the code for \method, along with instructions for using it on new data, available online.\footnote{\href{https://github.com/cbelth/ATP-morphology}{https://github.com/cbelth/ATP-morphology.}}

\section{Evaluation}
We evaluate the developmental plausibility of \method with respect to the order in which rules are acquired and its quantitative performance on realistic data. 
We further evaluate how the regularities \method discovers and its productions on Wug tests correspond to children's discoveries and productions.

\subsection{Data and Setup}

We briefly discuss the data that we use in our experiments, and name each dataset for reference.

\vspace{0.15cm}
\textbf{\childesGer} contains 442 German nominative singular/plural pairs of child-directed speech from the Leo \cite{behrens2006input} \childes corpus. Features encode gender, one of $\featureSpace =$ \{\texttt{feminine, masculine, neuter}\}, and frequency of words is known. We removed umlauts, as these follow a separate morphological process from suffixation. For developmental experiments, we sampled subsets of size 400, weighted by frequency, each sample modeling a child's 400 word vocabulary. The findings (reviewed earlier) that age-two children have productive plural morphology suggests that suffixes are learnable on vocabularies of this size or even smaller. While the \childes-derived words are used for training, we test on CELEX words. Morphological knowledge is generally acquired during childhood but must be able to generalize to other words in the lexicon. 

\vspace{0.15cm}
\textbf{\childesEng} was constructed from 6,539 word-inflection pairs extracted from child-directed English \cite{macwhinney2000childes}. There are 3,321 plural nouns (23 irregular), 1,494 past tense verbs (120 irregular), and 1,724 progressive verbs (the exceptionless \textit{-ing}). As reviewed earlier, children's vocabulary size during morphological learning is very modest. We thus log-binned words into 20 bins based on frequency and sampled 50 from each, simulating a child's vocabulary growth from 50 to 1K words. We repeated 100 times with different random seeds to simulate 100 children. Features are one of $\featureSpace = $ \{\texttt{progressive, past, plural}\}. 

\vspace{0.15cm}
\textbf{\celexEng} contains just the stem/past tense pairs from \childesEng, intersected with CELEX \cite{Baayen1996TheCL} for word frequency. For cross-validation, we formed 10 random 1000/100/200 train/dev/test splits, and for developmental experiments, we test on the top $n$ items, $n \in \{100, 200, 400, 600, 800, 1000\}$. To simulate children, we added random jitters between 0 and 5 to the frequencies. This has the Zipfian effect of scrambling which low-frequency items appear in each learner's training data. 

\subsubsection{Setup}
Throughout the experiments, we use orthography for German and IPA transcriptions for English, following \citet{McCurdy2020InflectingWT}'s use of orthography and \citet{Kirov2018RecurrentNN}'s use of phonological transcriptions. Token frequencies are  used to construct realistic datasets, but only type frequencies are used in learning and evaluation. 
We use paired $t$-tests at a 0.95 confidence level for statistical analysis. When testing \citet{Kirov2018RecurrentNN}'s ED model, we follow their setup, using an identical RNN implementation, trained for 100 epochs, with batch size 20. Both encoder and decoder RNNs are bidirectional LSTMs \cite{schuster1997bidirectional} with two layers, 100 hidden units, and a vector size of 300.

\subsection{Developmental Results}
\label{subsec:development}

\subsubsection{Order of Acquisition}
\label{subsubsec:order-of-ac}

\begin{figure}[t]
    \centering
    \includegraphics[width=\columnwidth]{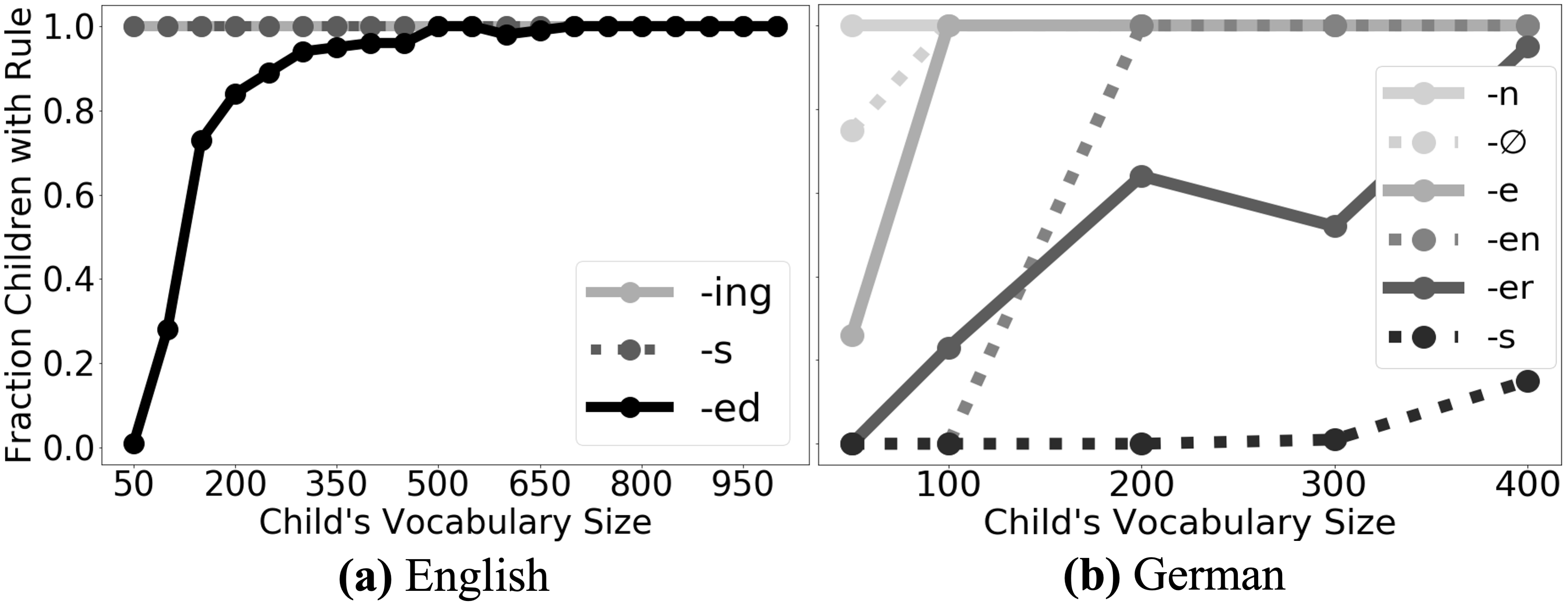}
    \caption{Fraction of ``children'' (\method runs on different data) who have learned each major suffixation rule. The order of acquisition closely follows child development. Legend order and shade of gray match acquisition order. All suffixes are consistently acquired, except German \textit{-s} ($\approx20$\%).}
    \label{fig:growth-curves}
\end{figure}

We study the order in which \method discovers productive processes in English and German to assess its consistency with children's developmental patterns on both English pluralization, verb past tense, and verb present participle (\childesEng) and German pluralization (\childesGer). 

Figure~\ref{fig:growth-curves} shows that \method acquires the \textit{-s} for pluralization (e.g., \texttt{book, books}) and \textit{-ing} for the verb present participle (e.g., \texttt{walk, walking}) on the smallest dataset (50 pairs). In contrast, the exception-laden \textit{-ed} of the past tense takes longer to overcome the irregulars, but is consistently acquired by around 500 words. These patterns align well with the order of morpheme acquisition by English-learning children \cite{Brown1973}.

On German pluralization, which has five primary suffixes (\textit{-(e)n}, \textit{-e}, -$\emptyset$, \textit{-er}, \textit{-s}; we separate \textit{-n} and \textit{-en}), \method acquires -n immediately (on 50 words), which closely follows 
``A" in \citet{Elsen2002}'s diary study, who learned \textit{-e(n)} words most quickly. By 100 words, -$\emptyset$ and -\emph{e} have been acquired; again matching 
``A," where the rate of learning of -$\emptyset$ and -\textit{e} words is virtually identical. Other longitudinal studies have also shown the very early acquisition of these suffixes \cite{Gawlitzek-Maiwald1994}. The \textit{-er} suffix is learned by 95\% of simulated children by 400 words. In \citet{Elsen2002}, \textit{-s} and \textit{-er} words were learned at similar rates, while other studies \cite{Kopcke1998, Bittner2000a, Szagun2001} also find that \textit{-s}, while productive, generally emerges later. In our data, \textit{-er} is attested 15 times, while \textit{-s} only 8 times;
\citet{Elsen2002} reports that A's \emph{s}-over-regularizations jump at 2;1, when her vocabulary contains around 50 -\emph{s} words, which may explain the difference.

\begin{figure}[t]
    \centering
    \includegraphics[width=\columnwidth]{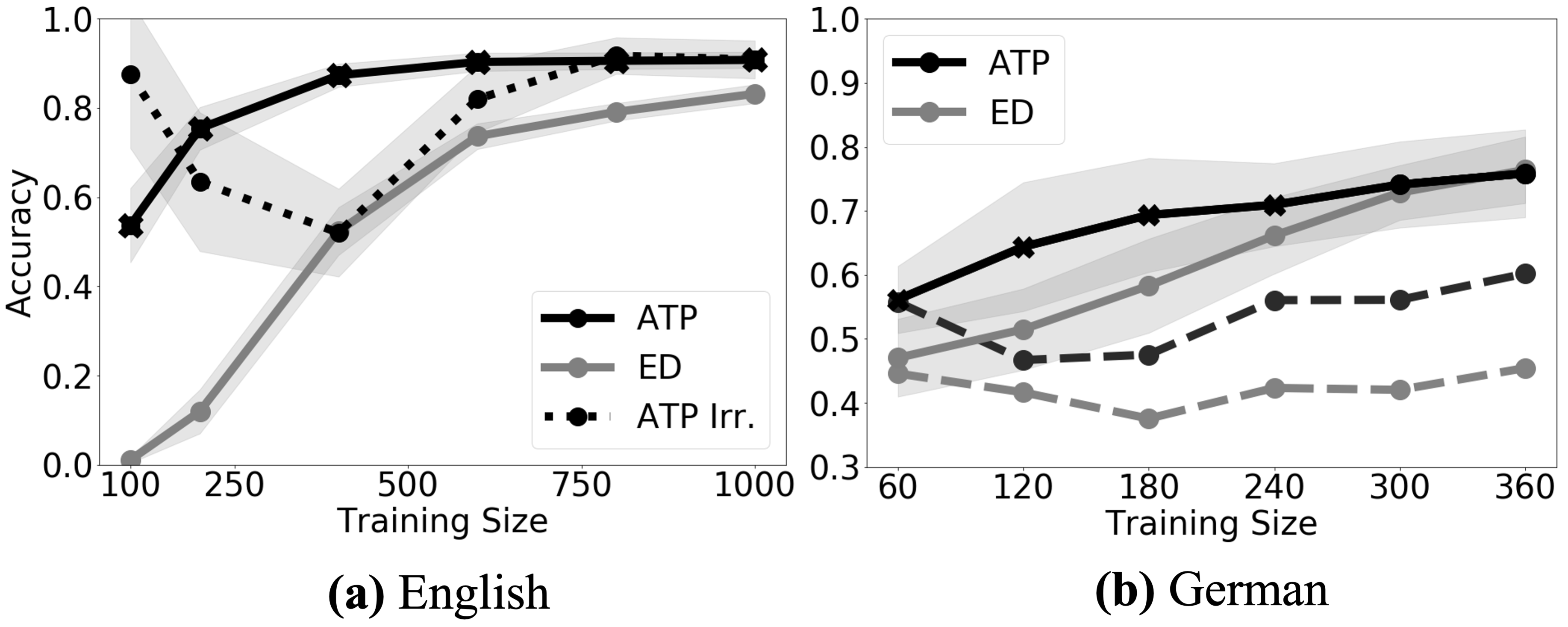}
    \caption{Accuracy on English and German vs. ED. \method outperforms ED by a stat.\ sig.\ amount on English and the first four training sizes for German. English accuracy on irregulars in the training data (dotted line) demonstrates the classic U-shape. The dashed line on German is the performance when the models are presented lemmas without gender.}
    \label{fig:performance-curves}
\end{figure}

\vspace{0.15cm}
\subsubsection{Performance}
\label{subsubsec:perf}

We evaluate quantitative performance on \celexEng and \childesGer in terms of how accurately the model generates inflections for held-out (\texttt{lemma}, \texttt{features}) pairs as the training data is progressively grown, simulating vocabulary growth. We compare to \citet{Kirov2018RecurrentNN}'s ED model. Results are in Fig.~\ref{fig:performance-curves}. 

On English past tense, \method outperforms ED by a statistically significant amount on all training sizes. Furthermore, when evaluated on irregulars that it has seen during training, \method over-applies learned rules before correcting, exhibiting 
the developmental regression \cite{Kuczaj1977, Marcus1992OverregularizationIL} that is a hallmark of the English past tense acquisition (dotted line). 
Before the model learns the productivity of -\textit{ed}, all training verbs are memorized (irregulars and regulars alike). Once the productive rule is learned, the model erases the memory of rule-following verbs, which no longer require storage. The two drops in irregular training accuracy correspond with the two largest jumps in test accuracy, demonstrating the acquisition of a rule (likely /-t/ and /-d/; see Fig.~\ref{fig:english-tree}). Throughout learning, all test errors were over-regularizations of rules (unless guessing occurred due to no rule having yet been acquired), matching the regularization vs. irregularization contrast in child acquisition \cite{berko1958child, MacWhinney1978, Xu1995}.

On German pluralization, \method outperforms ED by a statistically significant amount on the first four datasets (60-240 words) 
and is statistically indistinguishable after that point. Though the training data contains gender, the models can also be tested with novel nouns without gender information. 
Plural suffixation is not only conditioned on gender but also the phonological properties of nouns \cite{Wiese1996, zaretsky2015no}. \method always outperforms ED without gender, indicating that it is more capable of extracting the phonological regularities in the German system. 

In addition to its accuracy, \method runs in seconds on all dataset sizes, compared to minutes for ED.

We note that the setup of this evaluation differs in certain respects from that used in the neural modeling literature. In particular, each number in Fig.~\ref{fig:performance-curves} is the average \emph{test} accuracy of \emph{fully-trained} models on each of the training sets of that size (with the exception of the dotted U-shape line, as discussed above). In contrast, learning curves in works like \citet{Kirov2018RecurrentNN} report \emph{training} accuracy of \emph{partially-trained} models by epoch. Moreover, in such works, the model's final test-accuracy is an average only over different random initializations of the model, holding the training data constant. We chose our setup because our focus is on modeling development, and evaluating periodically on held-out test data is a more faithful measure of a model's developmental trajectory.

\subsection{Acquired Knowledge}
\label{subsec:knowledge}

\vspace{0.15cm}
\subsubsection{Discovered Rules}
\label{subsubsec:trees}

\begin{figure}[t]
    \centering
    \includegraphics[width=1.00\columnwidth]{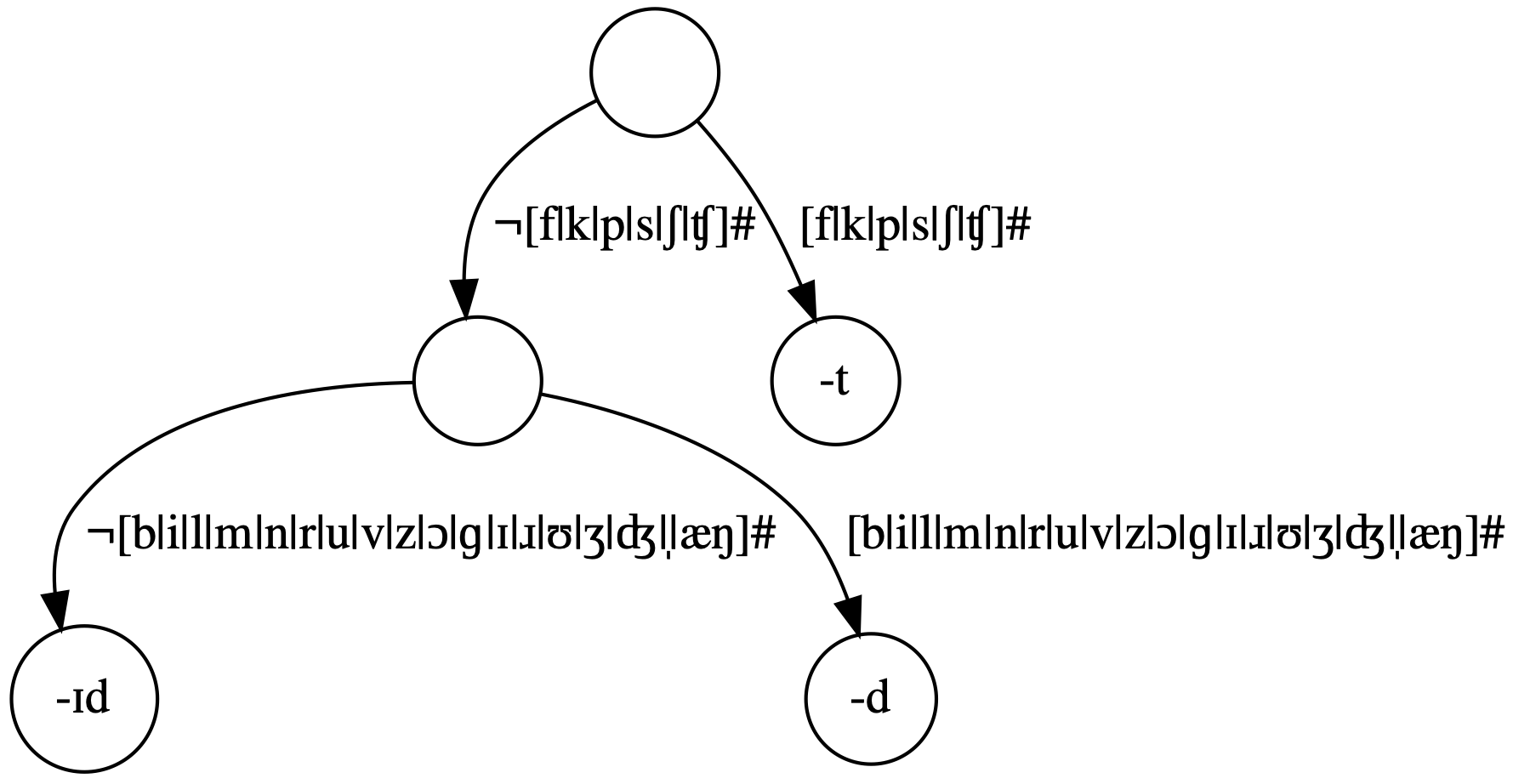}
    \caption{\method decision tree for \childesEng past tense. IPA symbols in brackets are separated by ``|'' to indicate that a lemma that ends in any of the listed endings follows (or does not follow, if preceded by ``$\lnot$'') the branch.}
    \label{fig:english-tree}
\end{figure}

\begin{figure}[t]
    \centering
    \includegraphics[width=1.00\columnwidth]{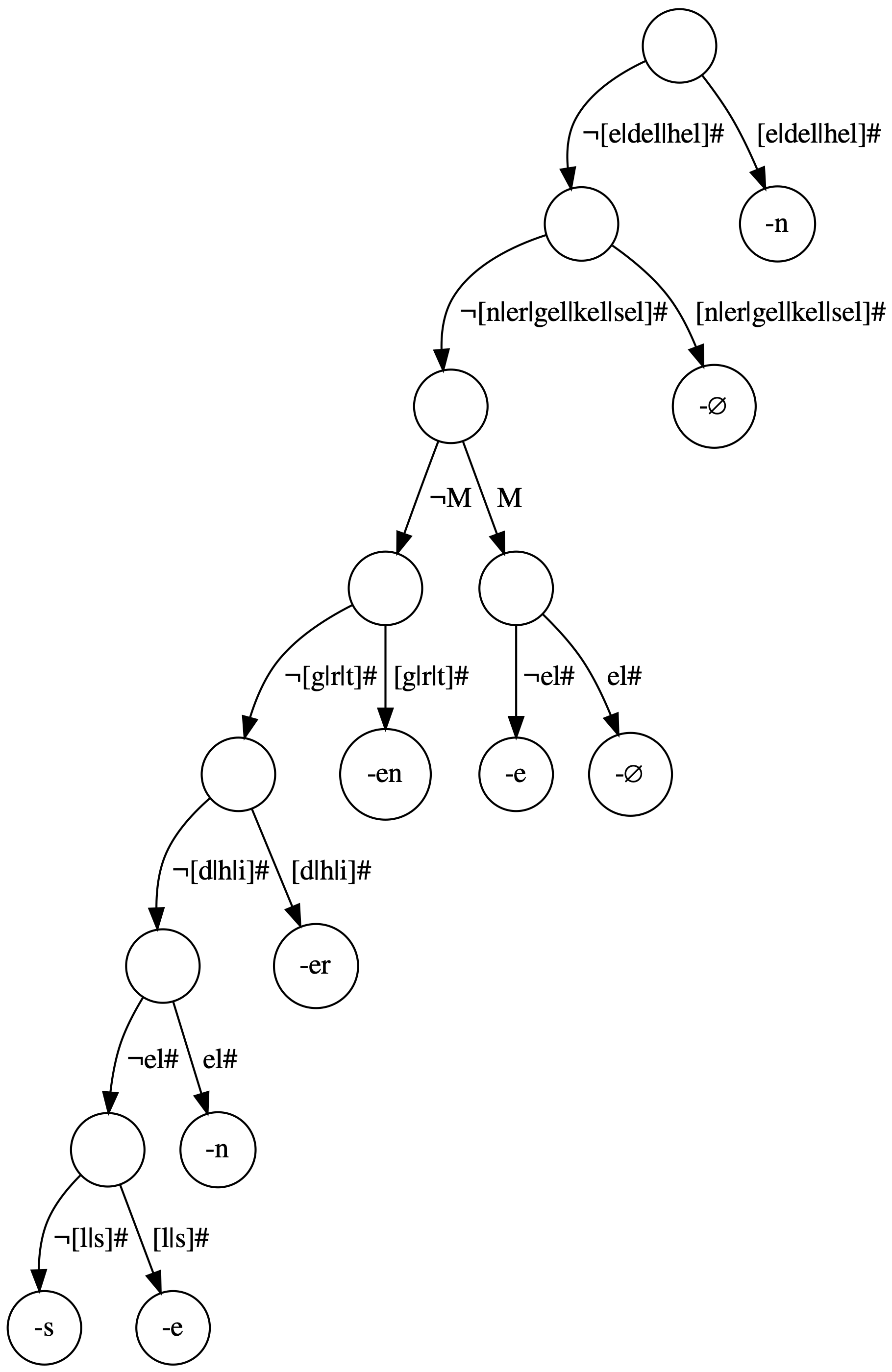}
    \caption{\method decision tree for \childesGer. ``M'' indicates nouns with masculine gender.}
    \label{fig:german-tree}
\end{figure}

ATP learns rules explicitly, represented in a decision tree.  The output trees in Figs.~\ref{fig:english-tree}-\ref{fig:german-tree} have transparent linguistic interpretations. 

The English tree (Fig.~\ref{fig:english-tree}) correctly characterizes the \textit{-ed} rule for English past tense: the first branch to the [-t] node splits on voiceless stem endings, the next branch to [-d] splits on voiced stem endings, and [-\ipaI d] captures the remaining instances. 

The German tree (Fig.~\ref{fig:german-tree}) captures all five primary suffixes (\textit{-(e)n}, \textit{-e}, -$\emptyset$, \textit{-er}, \textit{-s}). The \textit{-s} suffix---well-known to be the most idiosyncratic---is picked up at the deepest point in the tree, conforming to its role as the rule of last resort \cite{Marcus1995b}. 

This particular English tree was learned from 1000 words (one simulated child) of \celexEng, and the German tree was learned from the 442 German nouns in \childesGer.

\vspace{0.15cm}
\subsubsection{Wug Test Production}
\label{subsubsec:wug}
\begin{table}[t]
    \centering
    \small
    \caption{Correlations with human production results (bold $=$ stat. sig.). Training \method on just 400 words of child-directed speech yields higher correlation with human productions than training a NN on 22x times the data \protect\cite{McCurdy2020InflectingWT}.}
    \label{tab:wug}
    \resizebox{\columnwidth}{!}{
    \begin{tabular}{clcc|cccc}
		& & Neuter & & & Unknown &  \\
    	\toprule
    	& \%R & \%NR & $\rho$ & \%R & \%NR & $\rho$\\
    	\toprule
    	-(e)n & 0.17 & 0.04 & -0.26 & 0.19 & 0.23 & \textbf{0.43}\\
    	-e & 0.27 & 0.35 & -0.14 & 0.45 & 0.62 & 0.01\\
    	-$\emptyset$ & 0.11 & 0.0 & \textbf{0.55} & 0.07 & 0.00 & \textbf{0.55}\\
    	-er & 0.44 & 0.17 & \textbf{0.53} & 0.29 & 0.0 & \textbf{0.46}\\
    	-s & 0.01 & 0.44 & 0.3 & 0.01 & 0.15 & \textbf{0.64}\\
    	other & 0.00 & 0.00 &  & 0.00 & 0.00 & \\
      	\bottomrule
    \end{tabular}
}
\end{table}

The Wug test can be used to assess the morphological knowledge acquired by both humans and computational models.  \citet{Marcus1995b} carried out a Wug test study with 24 nonce words, divided into Rhyme (rhymes with familiar German words) and Non-Rhyme (unfamiliar) nonce words. The same stimuli were used by  \citet{zaretsky2015no} and \citet{McCurdy2020InflectingWT}.
\citet{McCurdy2020InflectingWT} compared the predictions of \citet{Kirov2018RecurrentNN}'s ED neural network model to human productions. We passed the same 24 nonce words to \method trained on \childesGer. This experiment follows \citet{McCurdy2020InflectingWT}: it computes (a) the fraction of productions for each suffix, divided into Rhyme (R) and Non-Rhyme (NR) and (b) Spearman's rank correlation ($\rho$) between the production probabilities of human and \method productions. To simulate multiple people, we ran \method on 500 samples of size 400 (by frequency) from \childesGer. As in \citet{McCurdy2020InflectingWT}, we treated each run as a model of a human, and computed the production probabilities for a particular word-suffix pair as the fraction of models that produced that suffix for that word. 

When presenting \method nonce words with unknown gender (right columns of Tab.~\ref{tab:wug}), \method correlates with human productions statistically significantly for all suffixes except \textit{-e}. This is in contrast to the ED model,  
which shows no correlation for any suffix \cite{McCurdy2020InflectingWT}. Moreover, \method was trained on a realistic 400 words of child-directed speech---over which the core inflectional morphology is learned---while ED was trained on 8.7K words, or 22x the data. Furthermore, both \citet{zaretsky2015no} and \citet{McCurdy2020InflectingWT} found that \textit{-(e)n} and \textit{-s} are used more frequently for non-rhyme words than rhyme words. \method again matches this behavior, and ED did not \cite{McCurdy2020InflectingWT}.

When presenting nonce words with unknown gender to \method, the correlation with human performance is higher than when presenting with neuter gender (left columns). The participants in \citet{McCurdy2020InflectingWT} were presented nonce words with the neuter determiner \emph{das}. However, the impact of this is unknown. As noted earlier \cite{Wiese1996}, the suffix choice is conditioned on both gender and phonology, and a conflict generally arises only in non-feminine gender (nearly all feminine nouns add -\textit{(e)n}). When they conflict, as in the test stimuli, human subjects may persist with a gender-conditioned rule, or they may eschew gender altogether and rely on phonological similarity to existing words \cite{zaretsky2015no}. Our model makes a testable prediction regarding this open question that can be pursued in future research.

\section{Discussion}
\method is not limited to morphology acquisition. Further research could investigate its use in learning phonology, syntax, or anything where linguistic generalizations are learned. Children's adeptness at language acquisition is a constant reminder of how much knowledge can be learned from tiny amounts of evidence. The decision trees that \method learns leave a step-by-step trace of what was learned and how. They thus provide explicit places to look for the steps children take in acquiring language. 

\section{Acknowledgments}
This work was supported by an NSF GRFP to CB. We thank the participants of the 2020 distributional learning seminar at the University of Pennsylvania for helpful discussion.

\bibliographystyle{apacite}

\setlength{\bibleftmargin}{.125in}
\setlength{\bibindent}{-\bibleftmargin}

\bibliography{ref}

\end{document}